\documentclass{bmvc2k}

\usepackage{multirow}
\usepackage{booktabs}
\usepackage{color}

\title{Continuous Event-Line Constraint for Closed-Form Velocity Initialization}

\addauthor{Xin Peng*}{pengxin1@shanghaitech.edu.cn}{123}
\addauthor{Wanting Xu*}{xuwt@shanghaitech.edu.cn}{123}
\addauthor{Jiaqi Yang}{yangjq@shanghaitech.edu.cn}{123}
\addauthor{Laurent Kneip}{lkneip@shanghaitech.edu.cn}{1}

\addinstitution{
 Mobile Perception Lab \\
 ShanghaiTech University \\
 Shanghai, China
}
\addinstitution{
  Shanghai Institute of Microsystem and Information Technology \\
  Chinese Academy of Sciences \\
  Shanghai, China
}
\addinstitution{
  University of Chinese Academy of Sciences \\
  Beijing, China
}

\runninghead{Student, Prof, Collaborator}{BMVC Author Guidelines}

\usepackage{enumitem}
\setenumerate[1]{itemsep=pt,partopsep=0pt,parsep=\parskip,topsep=5pt}
\setitemize[1]{itemsep=1pt,partopsep=0pt,parsep=\parskip,topsep=5pt}
\setdescription{itemsep=1pt,partopsep=0pt,parsep=\parskip,topsep=5pt}

\begin{document}

\maketitle

\begin{abstract}
Event cameras trigger events asynchronously and independently upon a sufficient change of the logarithmic brightness level. The neuromorphic sensor has several advantages over standard cameras including low latency, absence of motion blur, and high dynamic range. Event cameras are particularly well suited to sense motion dynamics in agile scenarios. We propose the continuous event-line constraint, which relies on a constant-velocity motion assumption as well as trifocal tensor geometry in order to express a relationship between line observations given by event clusters as well as first-order camera dynamics. Our core result is a closed-form solver for up-to-scale linear camera velocity {with known angular velocity}. Nonlinear optimization is adopted to improve the performance of the algorithm. The feasibility of the approach is demonstrated through a careful analysis on both simulated and real data.
\end{abstract}

\section{INTRODUCTION}
\label{sec:introduction}
{Event Cameras, such as the DVS~\cite{lichtsteiner2008128},} are bio-inspired visual sensors that differ substantially from traditional frame-based cameras. The pixels of an event camera operate asynchronously and trigger an event whenever there is sufficient change in the sensed logarithmic brightness level. 
More specifically, if the change of logarithmic brightness $L(\mathbf{x},t) \doteq \log I(\mathbf{x},t)$ at pixel $\mathbf{x} \doteq (x,y)^{\mathsf{T}}$ on the image plane surpasses a threshold $C$, the event camera will output a four-tuple signal $e = \left\{ x,\ y,\ t,\ s \right\}$ where $t$ is a timestamp and $s$ is a binary polarity indicating whether the brightness has increased or decreased. Therefore, each pixel has its own sampling rate and outputs data proportionally to the amount of motion between camera and scene and in dependence of the gradient of the visual input. An event camera does not produce images at a constant rate, but rather a stream of asynchronous, sparse events in a space-time volume with approximately microsecond time resolution.

Due to its exact nature, event cameras have several advantages over standard cameras including low latency ($\tilde1\mu$s), absence of motion blur, high dynamic range (140 dB vs 60 dB for traditional cameras~\cite{gallego2019focus}), and low power consumption. These beneficial properties enable an event camera to tackle vision tasks even in challenging conditions such as increased agility or low illumination conditions. 
One of the critical applications of an event camera is ego-motion estimation given that existing pipelines based on standard camera easily fail under high-speed motion or challenging illumination~\cite{mur2017orb, klein2007parallel}. However, the technical obstacle for event-based motion estimation is the fact that events are asynchronous and do not communicate absolute intensity information. Traditional motion estimation algorithms are therefore not appropriate and novel algorithms are needed.

Most of the state-of-the-art works in event-based motion estimation rely on learning-based approaches~\cite{kreiser2020chip}, filter-based methods~\cite{gallego2017event,kim2016real} and optimization methods~\cite{gallego2019focus,liu2021spatiotemporal,weikersdorfer2013simultaneous}. Learning-based approaches need a huge amount of data to train the network, and filter-based methods are computationally complex and often need an initial guess. Since most problems are nonlinear, the results of optimization methods highly depend on a good initial guess. \cite{peng2021globally,liu2020globally} 
{provide globally optimal solvers, which do not rely on good initial guess however they are computationally demanding and limited to homography environments. However, the methods are computationally demanding and limited to homography scenarios. Line features have already been used in event-based structure-from-motion frameworks. An example is given by Hough$^2$Map~\cite{Tschopp2021Hough2Map} which detects, tracks, and triangulates general lines. Other methods require highly artificial, black-and-white textures~\cite{mueggler2014event}.
Brändli et al. \cite{brandli2016elised} propose Event-Based Line Segment Detector (ELiSeD), which adopts the idea behind the LSD algorithm \cite{von2012lsd} to the event-based case.The method performs incremental event-based detection and tracking of lines in arbitrary scenes, but is not yet validated in the context of a full structure-from-motion framework. DVS sensors are often equipped with an Inertial Measurement Unit (IMU) (i.e. DAVIS 240~\cite{brandli2014240}), which is why researchers have also considered event-based visual-inertial odometry~\cite{mueggler2018continuous,le2020idol,vidal2018ultimate}
In particular, Le Gentil et al. \cite{le2020idol}introduce a line-based event-inertial odometry framework. In their event-based line tracking front-end, they draw concepts from \cite{brandli2016elised} and \cite{everding2018low}, and detect line segments as locally spatio-temporal planar patches. } 

{A critical concern in inertial odometry frameworks is given by bootstrapping. Our method aims at linear velocity initialization, which can neither be obtained directly from IMU, nor an event camera. Most existing event-inertial odometry algorithms pay little attention to the initialization question. Though using the assumption of known angular velocities, our work is the first to focus on linear velocity initialization and proposes a novel closed-form solver. It may be used to bootstrap other event-based inertial odometry frameworks, and in particular supports fusion at the level of velocities rather than positions. This has the advantage of requiring only single integration of inertial signals.} The main contributions are listed as follows:
\begin{itemize}
    \item We make use of trifocal tensor geometry~\cite{hartley2003multiple} to formulate the relationship between events, lines and the ego-motion of the event camera, the so-called Continuous Event-Line Constraint (CELC).
    \item To the best of our knowledge, this leads to the first closed-form translational velocity solver. It relies on the assumptions of known angular velocity and constant speed {resulting in} a linear constraint, and furthermore enables nonlinear optimization to improve performance.
    \item Important steps towards a DVS pendant of epipolar geometry and relative motion estimation for normal cameras.
\end{itemize}


The paper is organized as follows.
Section~\ref{sec:eventline_celc} reviews the general idea of trifocal tensor geometry with lines and introduces the continuous event-line constraint. In Section~\ref{sec:eventline_linear_solver}, we employ CELC for linear velocity estimation with a closed-form solution, and provide all implementation details. In  Section~\ref{sec:eventline_experiments}, we analyze and evaluate the proposed algorithm via both simulated and real experiments. Section~\ref{sec:eventline_conclusion} gives final remarks.

\section{Continuous Event-Line Constraint}
\label{sec:eventline_celc}

We give a review of the trifocal tensor with standard cameras. We then derive the continuous event-line constraint (CELC) for an event camera which indicates the relationship between events, lines and camera dynamics.

\subsection{Review of Trifocal Tensor}
\label{sec:eventline_Trifocal_Tensor}

The trifocal tensor plays an analogous role in three views to that played by the fundamental matrix in two~\cite{hartley2003multiple}. It encapsulates all the (projective) geometric relations between three views including the incidence relationship of three corresponding points, three corresponding lines, and point with line incidence relations. The line-point-line incidence relation is the most relevant for our work.

%
\begin{figure}[t]
    \centering
    \subfigure[]{
        \includegraphics[width=0.45\textwidth]{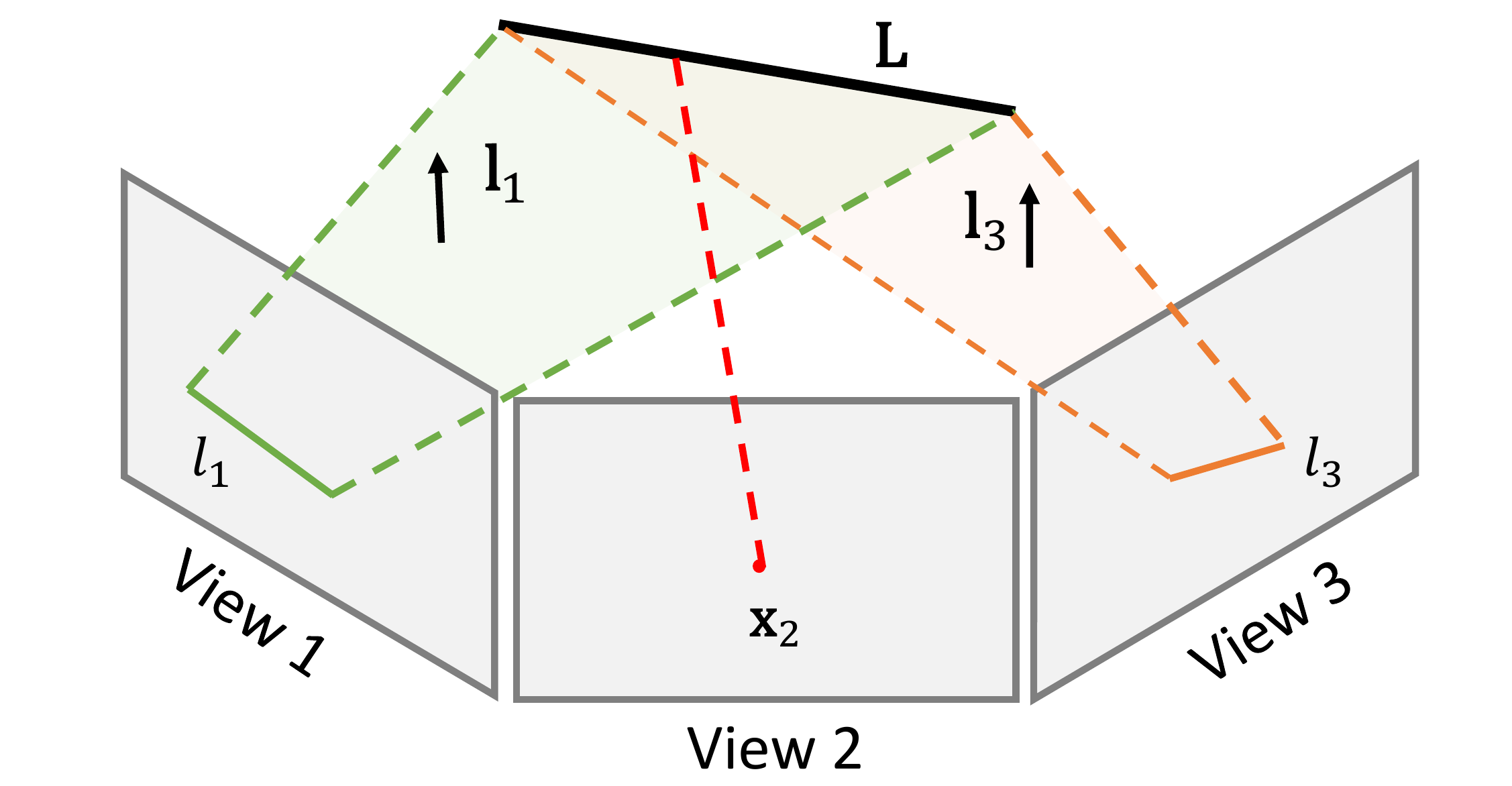}
        \label{fig:triview}
    }
    \subfigure[]{
        \includegraphics[width=0.45\textwidth]{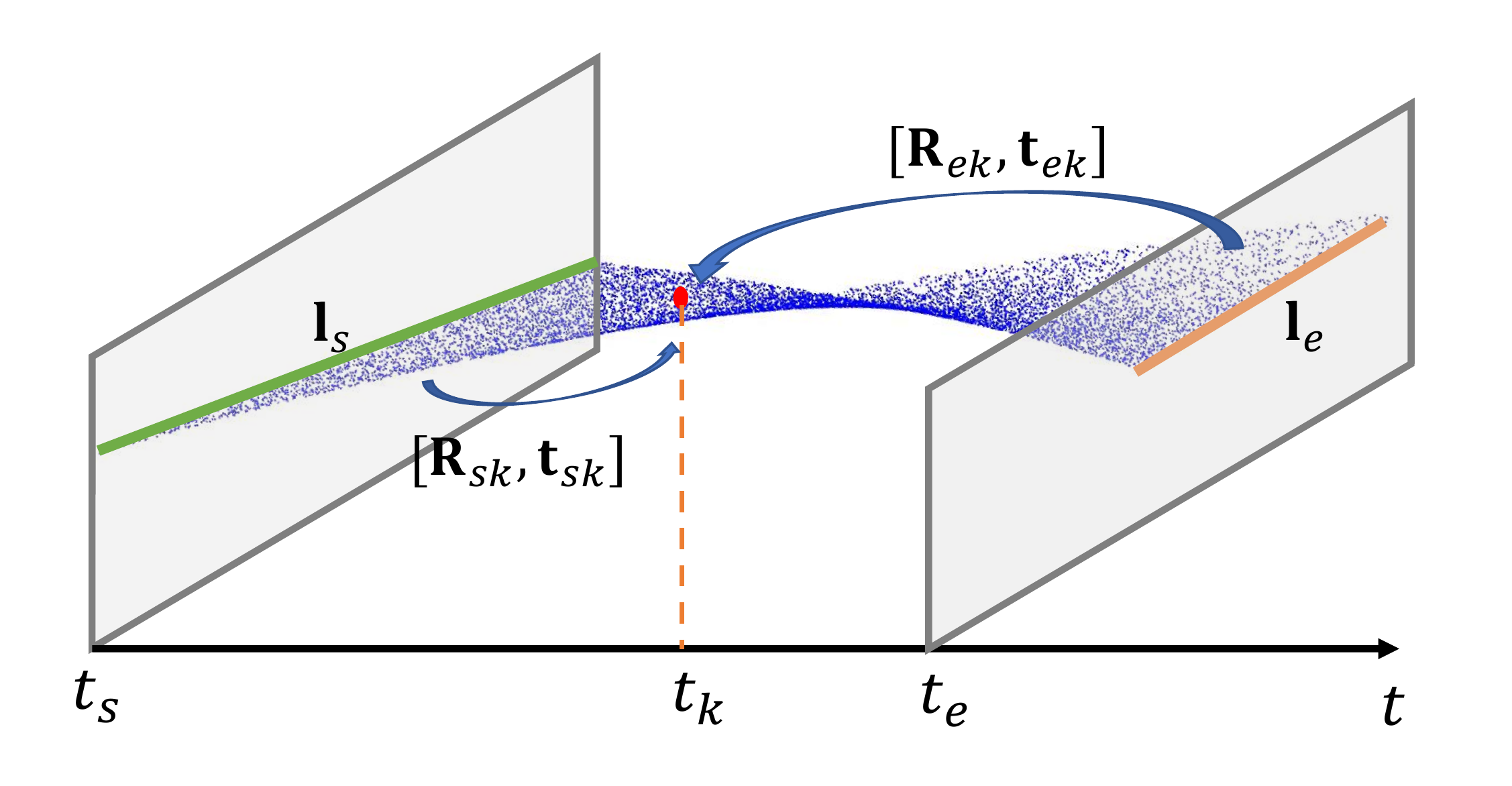}
        \label{fig:CELC}
    }
    \caption{Geometry of trifocal geometry.}
\end{figure}
Lets denote a 3D line $\mathbf{L}$, its two corresponding line projections ${l}_1$ and ${l}_3$ in views 1 and 3, respectively, and an image point $\mathbf{x}_2$ in view 2 which is the projection of a 3D point on $\mathbf{L}$. The geometry is illustrated in Fig.~\ref{fig:triview}. Let's define the second view as the reference view. We furthermore define $ [\mathbf{R}_{12}| \mathbf{t}_{12}]$ as the Euclidean transformation parameters from view 2 to view 1, and $[\mathbf{R}_{32}| \mathbf{t}_{32}]$ as the Euclidean transformation parameters from view 2 to view 3. 
In the calibrated case, the line-point–line incidence relation is given by
\begin{equation}
    \mathbf{f}_2^{\mathsf{T}} (\mathbf{l}_1^{\mathsf{T}}[\mathbf{T}_1,\mathbf{T}_2,\mathbf{T}_3]\mathbf{l}_3) = 0,
    \label{eqn:trifocal}
\end{equation} 
where $\mathbf{f}_2$ is the bearing vector corresponding to pixel $\mathbf{x}_2$ in view 2, and $\mathbf{l}_i = \mathbf{K}^{\mathsf{T}} {l}_i$ with $i=1,3$ are the normal vectors of the planes crossing $\mathbf{L}$ and the camera centers of views 1 and 3, respectively.  $[\mathbf{T}_1,\mathbf{T}_2,\mathbf{T}_3]$ defines the trifocal tensor, and the formulation for the $3\times 3$ matrices $\mathbf{T}_i$ is given by
\begin{equation}
    \mathbf{T}_i = \mathbf{r}_i^{12}\mathbf{t}_{32}^{\mathsf{T}} - \mathbf{t}_{12} \mathbf{r}_i^{32 \mathsf{T}},\  i=1,2,3,
    \label{eqn:trifocal_tensor}
\end{equation}
where $\mathbf{r}_i^{12}$ denotes the $i$-th column of $\mathbf{R}_{12}$, and $\mathbf{r}_i^{32}$ the $i$-th column of $\mathbf{R}_{32}$. For further details on the trifocal tensor including its derivation, the reader is kindly referred to Chapter 15 of \cite{hartley2003multiple}.

\subsection{Continuous Event-Line Constraint -- CELC}

Our task consists of event camera ego-motion estimation. Our assumption is that events are mostly triggered by the reprojection of sharp appearance and occlusion edges, which---for the sake of a simplified derivation---are furthermore assumed to be straight in 3D. Note that this assumption may not be limiting, as {man-made} environments often present themselves in a form where the majority of such edges are indeed straight.
%

We detect continuous line projections from event streams and figure out the relation between those events, the underlying 3D lines, and dynamic motion parameters by using the aforementioned trifocal tensor relations. The continuous set of events triggered by the projection of a straight 3D line $\mathbf{L}$ under motion forms a cluster of events $\mathcal{E}$ in a twisted manifold-like shape. As denoted by the green and red lines in Figure~\ref{fig:CELC}, let the two lines ${l}_1$ and ${l}_3$ represent the reprojection of the 3D line $\mathbf{L}$ at timestamps $t_s$ and $t_e$, respectively. As introduced in Sec.~\ref{sec:eventline_Trifocal_Tensor}, for an event $e_k \in \mathcal{E}$, ${l}_1$ and ${l}_3$ must then satisfy the trifocal relation (\ref{eqn:trifocal}). Here, let's define $[\mathbf{T}_{1}^k,\mathbf{T}_2^k,\mathbf{T}_3^k]$ as the trifocal tensor for the $k$-th event. The trifocal tensor is constructed by the transformation of the camera from $t_k$ to $t_s$ and the transformation from $t_k$ to $t_e$. The trifocal tensor will be different for each individual event. Note that rather than introducing an individual rotation and translation for each event---which would introduce too many unknowns---, we make use of a locally constant velocity assumption and parameterize the relative translation and rotation as a continuous time function of the linear velocity $\mathbf{v}$ and angular velocity $\boldsymbol{\omega}$. Hence, the rotation $ \mathbf{R}_{sk}$ from time $t_k$ to time $t_s$ can be represented by the continuous time function
\begin{eqnarray}
    \mathbf{R}_{sk} &=& \exp(\hat{\boldsymbol{\omega}} (t_k - t_s)) \nonumber\\
    &=& \cos(\theta) \mathbf{I} + (1-\cos(\theta))\mathbf{a}\mathbf{a}^{\mathsf{T}} + \sin(\theta)\hat{\mathbf{a}},
    \label{eqn:angular_velocity2rotaion}
\end{eqnarray}
%
where $(\mathbf{a}, \theta)$ are the axis-angle parameters of the rotation $\mathbf{R}_{sk}$. The translation $\mathbf{t}_{sk}$ from time $t_k$ to time $t_s$ is given by
\begin{eqnarray}
    \mathbf{t}_{sk} &=& \mathbf{J}_{sk} \mathbf{v} (t_k - t_s), \nonumber \\
    \text{where }\ \mathbf{J}_{sk} &=& \frac{\sin(\theta)}{\theta} \mathbf{I} + (1-\frac{\sin(\theta)}{\theta})\mathbf{a}\mathbf{a}^{\mathsf{T}} + \frac{1-\cos{\theta}}{\theta}\hat{\mathbf{a}}.
    \label{eqn:velocity2translation}
\end{eqnarray}
By replacing $t_s$ with $t_e$ in equations (\ref{eqn:angular_velocity2rotaion}) and (\ref{eqn:velocity2translation}), we can obtain $ \mathbf{R}_{ek}$ and $\mathbf{t}_{ek}$. Based on (\ref{eqn:angular_velocity2rotaion}) and (\ref{eqn:velocity2translation}), we finally obtain the continuous time formulation of the trifocal tensor
%
\begin{eqnarray}
    \mathbf{T}_i &=& \mathbf{r}_i^{sk}\mathbf{t}_{ek}^{\mathsf{T}} - \mathbf{t}_{sk} \mathbf{r}_i^{ek \mathsf{T}},\  i=1,2,3, \nonumber \\
    &=& \mathbf{r}_i^{sk}(\mathbf{J}_{ek} \mathbf{v} (t_{k} - t_{e}))^{\mathsf{T}} - (t_{k} - t_{s}) \mathbf{J}_{sk} \mathbf{v} \mathbf{r}_i^{ek \mathsf{T}} .
    \label{eqn:eventline_trifocaltensor}
\end{eqnarray}
Using {equation (\ref{eqn:eventline_trifocaltensor})} and applying simple matrix multiplication, we obtain
\begin{eqnarray}
    \mathbf{l}_1^\mathsf{T} \mathbf{T}_i \mathbf{l}_3 &=& \mathbf{l}_1^\mathsf{T} \left[ (t_{k} - t_{e}) \mathbf{r}_i^{sk} \mathbf{v}^\mathsf{T}  \mathbf{J}_{ek}^\mathsf{T}  - (t_{k} - t_{s}) \mathbf{J}_{sk} \mathbf{v} \mathbf{r}_i^{ek \mathsf{T}} \right] \mathbf{l}_3 \nonumber \\
    &=&  (t_{k} - t_{e}) \mathbf{l}_1^\mathsf{T} \mathbf{r}_i^{sk} \mathbf{l}_3^\mathsf{T} \mathbf{J}_{ek} \mathbf{v}  - (t_{k} - t_{s})  \mathbf{l}_3^{\mathsf{T}} \mathbf{r}_i^{ek} \mathbf{l}_1^\mathsf{T} \mathbf{J}_{sk} \mathbf{v}.
    \label{eqn:eventline_temperal}
\end{eqnarray}
%
The continuous event-line constraint (CELC) for the $k$-th event is obtained by combining (\ref{eqn:eventline_temperal}) and (\ref{eqn:trifocal}), thus resulting in
\begin{equation}
    \mathbf{f}_k^{\mathsf{T}} \mathbf{B}_k \mathbf{v} = 0,
    \label{eqn:eventline_CELC}
\end{equation}
where
\begin{equation}
    \mathbf{B}_k = \left[ \begin{matrix}
    (t_{k} - t_{e}) \mathbf{l}_1^\mathsf{T} \mathbf{r}_1^{sk} \mathbf{l}_3^\mathsf{T} \mathbf{J}_{ek}  - (t_{k} - t_{s}) \mathbf{l}_3^{\mathsf{T}} \mathbf{r}_1^{ek} \mathbf{l}_1^\mathsf{T} \mathbf{J}_{sk} \\
    (t_{k} - t_{e}) \mathbf{l}_1^\mathsf{T} \mathbf{r}_2^{sk} \mathbf{l}_3^\mathsf{T} \mathbf{J}_{ek}  - (t_{k} - t_{s}) \mathbf{l}_3^{\mathsf{T}} \mathbf{r}_2^{ek} \mathbf{l}_1^\mathsf{T} \mathbf{J}_{sk} \\
    (t_{k} - t_{e}) \mathbf{l}_1^\mathsf{T} \mathbf{r}_3^{sk} \mathbf{l}_3^\mathsf{T} \mathbf{J}_{ek}  - (t_{k} - t_{s}) \mathbf{l}_3^{\mathsf{T}} \mathbf{r}_3^{ek} \mathbf{l}_1^\mathsf{T} \mathbf{J}_{sk} \\
    \end{matrix}
    \right] .
\end{equation}

The incidence relation expresses the intrinsic relationship between events generated by a 3D line and first order camera dynamics. Considering the transformation of lines and trifocal tensor geometry, $\tilde{\mathbf{l}}_2 = \mathbf{B}_k \mathbf{v}$ represents the projected line in view $k$ generated by reference lines $\mathbf{l}_1$ and $\mathbf{l}_3$ and motion dynamics. Any event $\mathbf{e}_k$ triggered by the same line should lie on $\tilde{\mathbf{l}}_2$, i.e. $\mathbf{f}_k^{\mathsf{T}} \tilde{\mathbf{l}}_2 = 0$.
As scale is unobservable in the monocular setting, the unknown motion parameters $\boldsymbol{\omega}$ and $\mathbf{v}$ actually make up for only 5-DoF. However, equation (\ref{eqn:angular_velocity2rotaion}) and (\ref{eqn:velocity2translation}) are nonlinear with respect to $\boldsymbol{\omega}$ and $\mathbf{v}$, which makes it hard to simultaneously figure out angular velocity and linear velocity. In the continuation, we therefore consider the case where angular velocities are given by an Inertial Measurement Unit (IMU).

\section{Closed-form Velocity Initialization}
\label{sec:eventline_linear_solver}



Typically, a DVS sensor such as the DAVIS346, integrates an event camera and an IMU which provides angular velocity and acceleration. With the help of the prior known angular velocity, the nonlinear 5 DoF motion estimation problem is reduced to a 2 DoF problem: translational velocity estimation. 
The closed-form speed initialization algorithm proceeds in four steps. The first step consists of event clustering. Next, for each cluster we extract the lines $\mathbf{l}_1$ and $\mathbf{l}_3$ {by using a small time interval of events} at the beginning and the end of each cluster. Finally, using (\ref{eqn:eventline_CELC}), we propose a linear closed-form speed solver for the linear velocity $\mathbf{v}$. The fourth and final step consists of nonlinear optimization improving the estimation result.


\subsection{Line Clustering and Extraction}
\label{sec:line_extraction}
%
%

We adopt a strategy similar to the one leveraged in~\cite{le2020idol,everding2018low}, which considers events as a 3D point cloud in the space-time volume. The coordinates are given by the pixel position of the event and the timestamp, i.e. $e_i = [x_i, y_i, t_i/c]$. To balance the magnitude of the image coordinates and the timestamp of an event, the latter is normalized by a constant $c$ whose value is chosen according to the average level of texture in the scene. The time span over which event clusters are formed is dynamically defined by considering a fixed number of $N$ events. Events generated by the same line will approximately form a local plane in the 3D space-time volume of the event stream. Hence---ignoring the influence of rotational velocities, clustering events generated by the same line in 3D roughly amounts to plane clustering in a 3D point cloud.
We employ the open-source C++ library \textit{Cilantro}~\cite{zampogiannis2018cilantro} to implement the clustering procedure, which operates in a region growing fashion inspired by connected component segmentation. For more details, please refer to~\cite{le2020idol}.
 

For each event cluster $\mathcal{E}_j$ in which events are sorted with increasing timestamps, we utilize the first and last {0.005s intervals of events} to extract the lines $l_{1j}$ and $l_{3j}$. We use \textit{cv::fitLine} from \textit{OpenCV} \cite{opencv_library} to extract the lines, and the algorithm is based on an M-estimator that iteratively fits the line using a weighted least-squares algorithm. We choose the Huber norm strategy (\citep{huber2004robust}, page 43). Note that our algorithm uses variable timestamps around which $l_{1j}$ and $l_{3j}$ are fitted, which enables us to slide the 0.005s intervals towards the center of the entire cluster interval in case of insufficient events.

\subsection{Linear Velocity Solver}

With known angular velocity, the CELC (\ref{eqn:eventline_CELC}) becomes linear in the translational velocity. It is furthermore easy to concatenate all linear constraints for all events of all clusters into one constraint. Given $M$ event clusters $\mathcal{E}_j$ where $j=1,2...,M$ and the corresponding extracted lines $l_{1j}$ and $l_{3j}$, the constraints from each event cluster with $N_{j}$ events can be stacked into the single linear problem

\begin{equation}
    \left[ \begin{matrix}
    \mathbf{B}_{11}^{\mathsf{T}} \mathbf{f}_{11}~
    \dots ~
    \mathbf{B}_{kj}^{\mathsf{T}} \mathbf{f}_{kj}  ~
    \dots ~
    \mathbf{B}_{N_M M}^{\mathsf{T}} \mathbf{f}_{N_M M}
    \end{matrix} \right]^\mathsf{T} \mathbf{v} = \mathbf{A} \mathbf{v} = 0.
    \label{eqn:linear_constraint}
\end{equation}

%
    
%
$\mathbf{A}$ can be computed from the known angular velocity, the extracted lines $l_{1j}$ and $l_{3j}$, and all measured events. This linear problem could be solved using $\mathbf{A}^{\mathsf{T}} \mathbf{A} \mathbf{v} = 0$ via SVD. However, given that least square estimation methods lack robustness \citep{huber2004robust}\citep{andrews1974robust}, we choose to use another robust M-estimator with a Huber norm \citep{huber2004robust}, and employ iteratively re-weighted least-squares fitting for the nullspace extraction. As the number of events $N$ is very large (about 100,000 in our real data experiments), and $\mathbf{A}$ is an $N \times 3$ matrix, we improve efficiency by randomly selecting 1000 samples out of the $N$ to perform the M-estimation. Further details of our implementation can be found in the \textit{Definition} part of \textit{Chapter 1.3} in \citep{ruckstuhl2014robust}. 



\subsection{Degenerated Case}

{
Note that our linear solver cannot always determine a unique solution. It is obvious that if the motion of the camera is a pure rotation---meaning that $\mathbf{v} = \mathbf{0}$---solving our linear equation~\ref{eqn:linear_constraint} via SVD will not be possible. Another degenerate case exists if the camera moves along a straight line without rotation. It is obvious that in this case any translational velocity component along the direction of the 3D line $\mathbf{L} = \mathbf{l}_{1} \times \mathbf{l}_{3}$ will not contribute to any appearance changes in the image, and therefore also no events. Hence, the 3D line direction needs to lie in the null space of the matrix $\mathbf{A}$, which means $\mathbf{A} (\mathbf{l}_{1} \times \mathbf{l}_{3}) = 0$. Moreover, $\mathbf{A} \mathbf{v} = \mathbf{A} (\mathbf{v}_1 + \mathbf{l}_{1} \times \mathbf{l}_{3}) = \mathbf{A} \mathbf{v}_1$. There exists an unobservable direction for the translational velocity given by the direction of the 3D line. In the case of linear motion, the unobservable direction also exists when there are multiple lines but all of them are parallel.}


\subsection{Nonlinear Optimization}
\label{sec:nonliear_optimization}

 
In real cases, events are affected by both spatial and temporal noise as well as outliers in the form of salt and pepper noise. This may lead to errors in the event clustering and the line extraction. The linear solver introduced in the previous section therefore only offers an initial guess, which may be further refined by a maximum likelihood estimation to make the estimation more accurate and robust.

The objective is to minimize the geometric distance between the reprojected 3D line and the events. The cost function is given by
\begin{equation}
    \min_{\mathbf{v}} \sum^{M}_{j=1} \sum^{N_j}_{k=1} d(\tilde{\mathbf{l}}_{2kj}, \mathbf{f}_{kj})^2, 
\end{equation}
where $d(\cdot)$ represents the distance function, and $\tilde{\mathbf{l}}_{2kj}$ is the reprojected line in the image plane at timestamp $t_{kj}$. One way to conveniently represent the 3D line during the nonlinear optimization is by its projections $\mathbf{l}_{1j}$ and $\mathbf{l}_{2j}$ in their relevant views. Given a candidate linear velocity, one can then again compute the trifocal tensor, and furthermore easily derive the location of the projected 3D line at the time of the corresponding event using the line transfer equation $\tilde{\mathbf{l}}_{2kj} = \mathbf{B}_{kj} \mathbf{v}$.

The entire objective is minimized using a trust-region based method (e.g. Levenberg-Marquardt) and implemented with \textit{Ceres}~\citep{ceres-solver} {using a robust cost function (Huber norm)}.


\section{Experiments}
\label{sec:eventline_experiments}

We analyze the proposed closed-form algorithm both in simulation and on real data. 
We use the Euclidean distance $\epsilon$ and the cosine distance $\phi$ between the estimated results and ground truth as a metric to evaluate the accuracy of the estimated results. They are given as follows:
{
\begin{equation}
	\epsilon = \|\mathbf{v}_{\text{gt}} - \mathbf{v}_{\text{est}} \|_2, \ \phi = \arccos( \mathbf{v}_{\text{gt}}^{\mathsf{T}} \mathbf{v}_{\text{est}} ),
\end{equation}
}
where $\mathbf{v}_{\text{gt}}$ and $\mathbf{v}_{\text{est}}$ are the ground truth and estimated linear velocities, respectively. 

\subsection{Simulation}
We start by evaluating the performance of the proposed approach over synthetic data. To generate synthetic data, we randomly generate line segments in 3D space within a volume of $x=[-2,2]$ m, $y=[-2,2]$ m, and $z=[3,6]$ m. With given angular and linear velocities, an event is generated by randomly choosing a 3D point on a line and projecting it into an image plane with the camera pose sampled by a random timestamp within a given time interval. In our experiments, we set the angular velocity of the camera as $\boldsymbol{\omega} = [0,\ 0,\ 2]$ rad/s, and the linear velocity as $\mathbf{v} = [1,\ 2,\ 0]$ m/s. We generate events from 5 lines within a time interval of 0.5 s. We disturb the pixel location of each generated event by zero-mean Gaussian noise with a standard deviation of 2 pixels. We also add Gaussian noise of $\mathcal{N}(0,2)$ pixels to the ground-truth endpoints of each starting and ending line pair $l_{1j}$ and $l_{3j}$.

\begin{figure}[t]
    \centering
    \subfigure[]{
        \includegraphics[width = 0.3\textwidth]{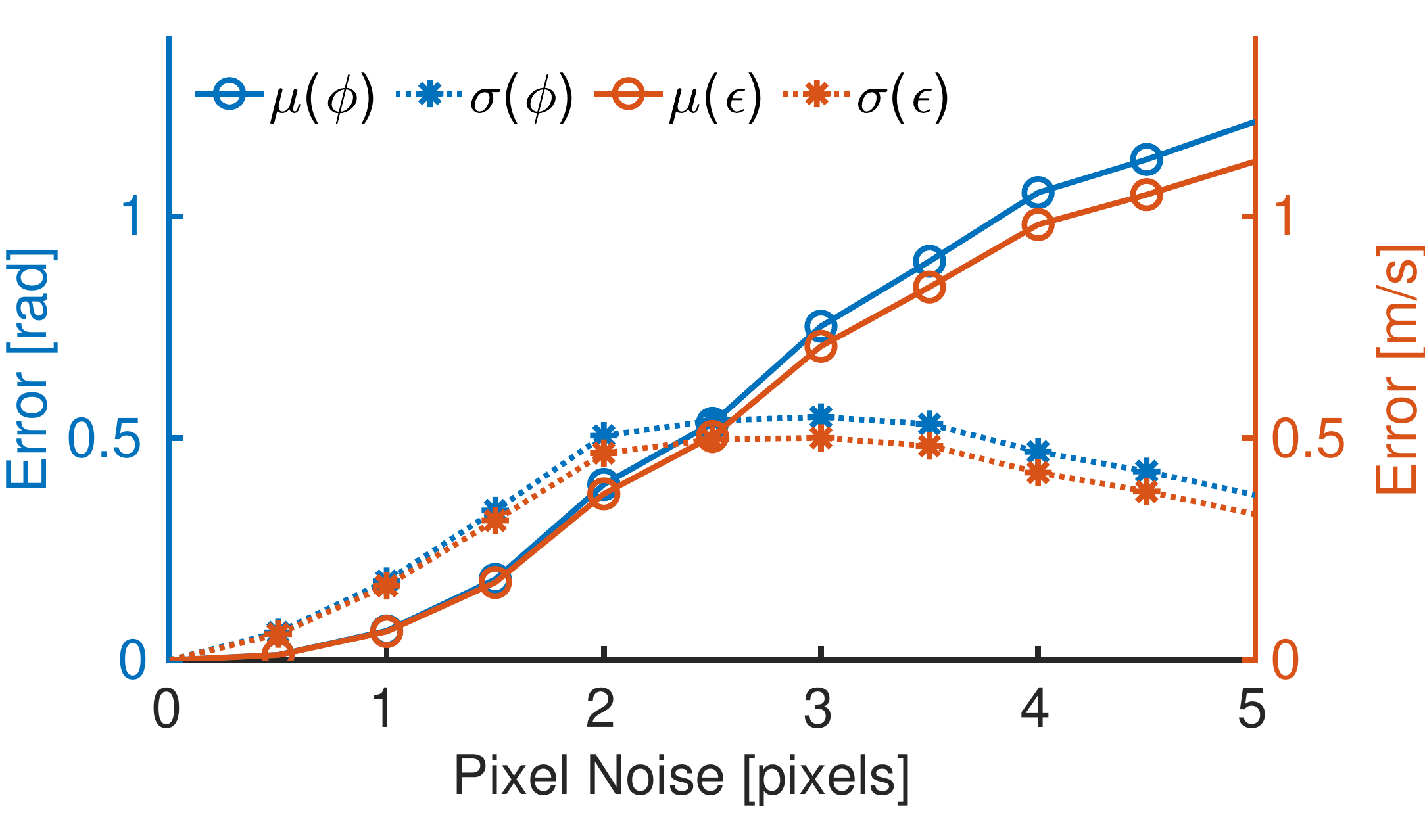}
        \label{fig:eventline_eventNoise}
    }
    \subfigure[]{
    \includegraphics[width = 0.3\textwidth]{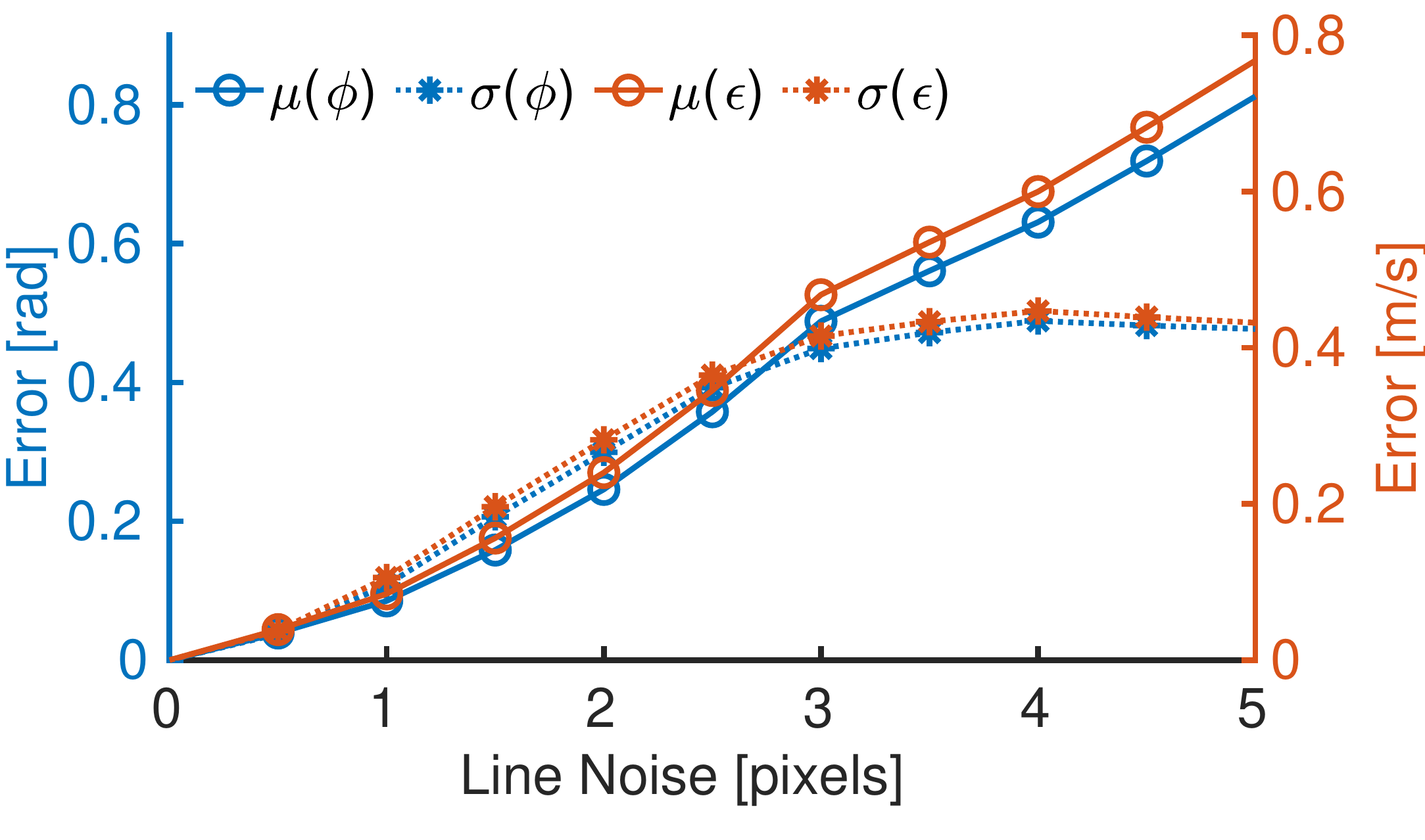}
    \label{fig:eventline_lineNoise}
    }
    \subfigure[]{
    \includegraphics[width = 0.3\textwidth, height = 0.18\textwidth]{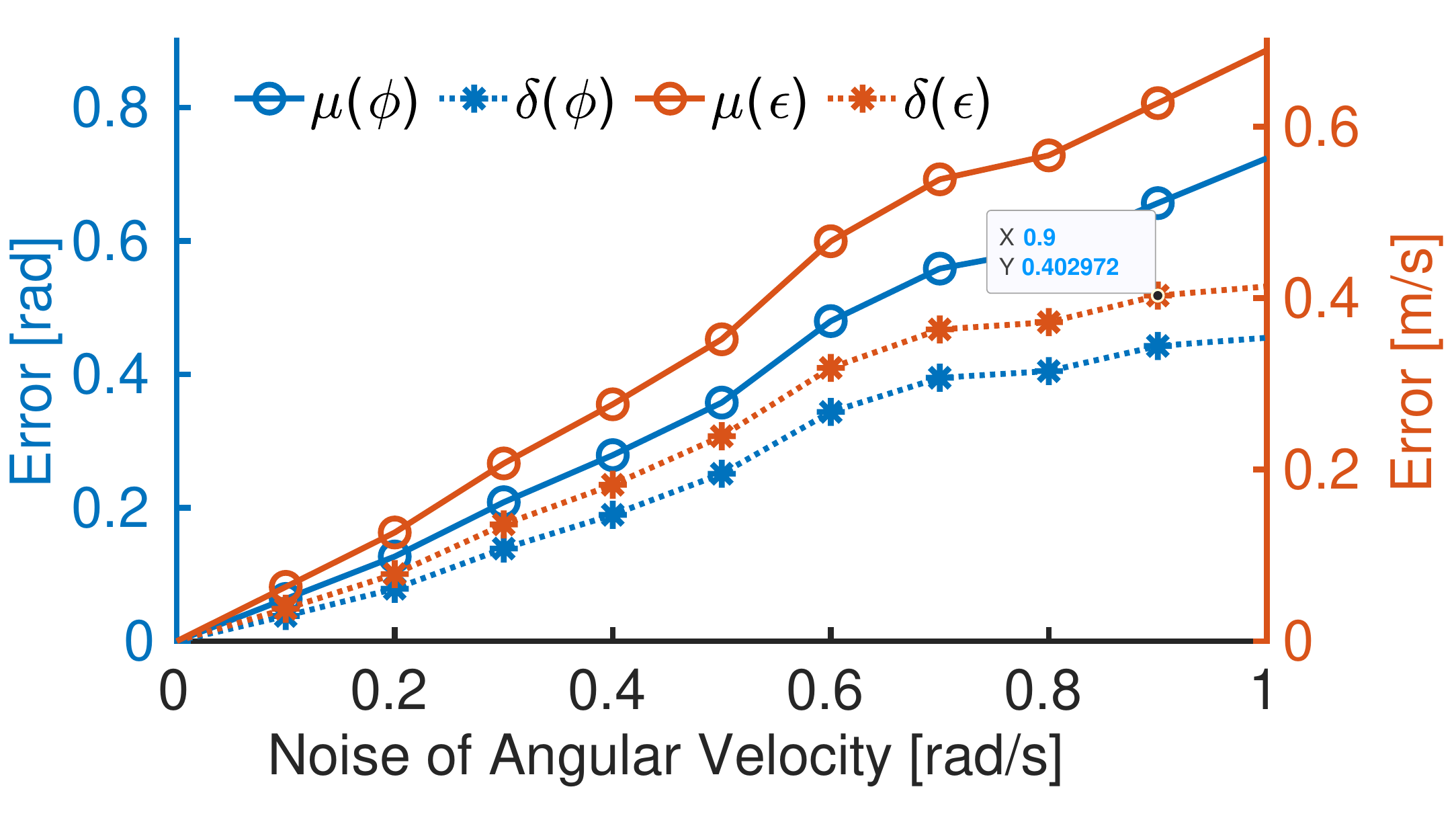}
    \label{fig:eventline_angularNoise}
    }
    \vspace{-0.2cm}
    \caption{Noise analysis. (a) shows the errors of the solver with increasing event location disturbance. (b) illustrates the error of our solver with different levels of noise added to the end-points of the fitted lines $l_{1j}$ and $l_{3j}$. (c) displays the errors of the solver with increasing angular velocity disturbance. The error generally increases with noise.}
    \label{fig:eventline_noise}
\end{figure}

\begin{figure}[t]
    \centering
    \subfigure[]{
        \includegraphics[width = 0.3\textwidth]{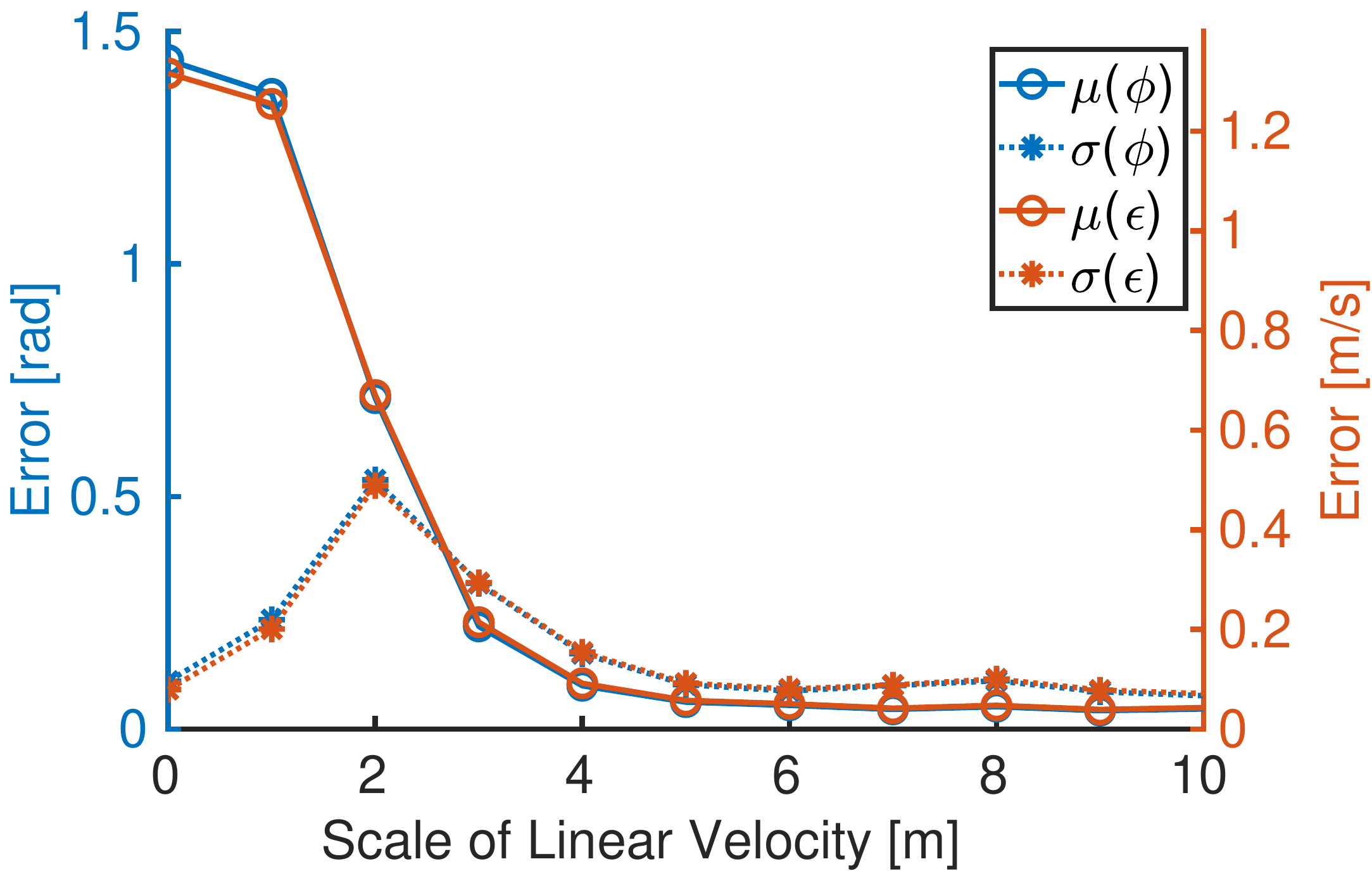}
        \label{fig:eventline_speed}
    }
    \subfigure[]{
    \includegraphics[width = 0.3\textwidth]{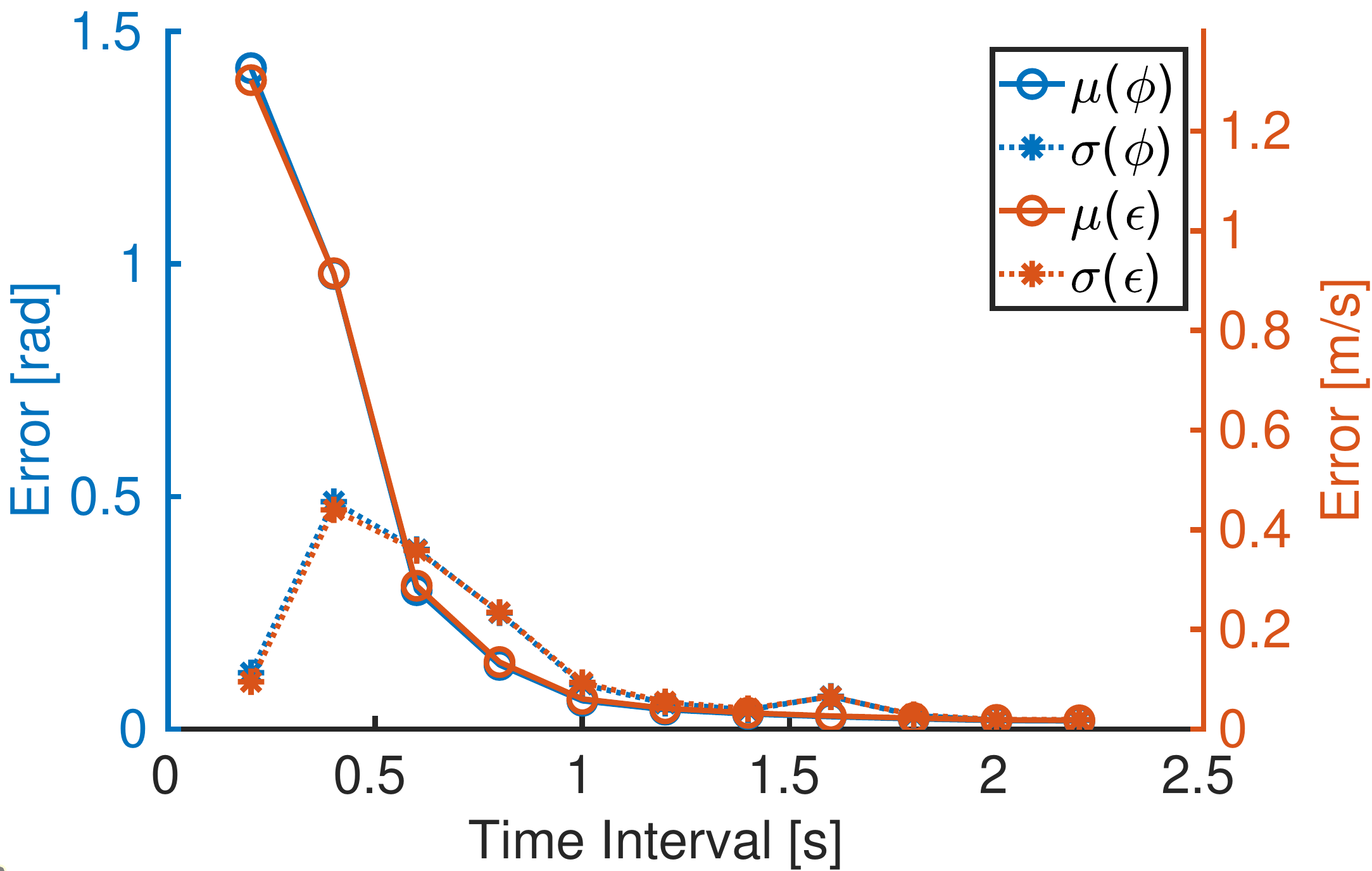}
    \label{fig:eventline_timeInterval}
    }
    \subfigure[]{
        \includegraphics[width = 0.3\textwidth]{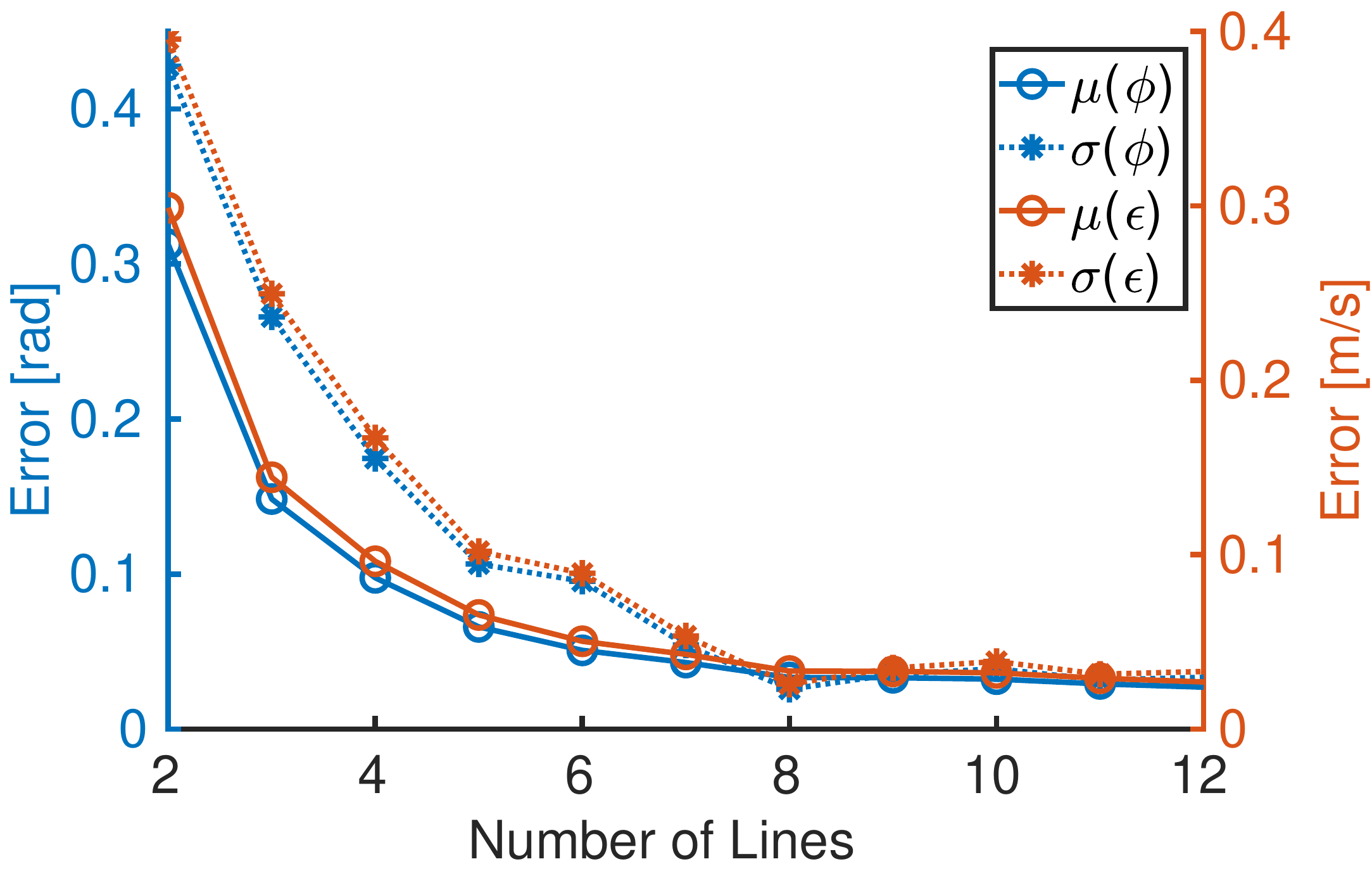}
        \label{fig:eventline_lineNum}
    }
    \vspace{-0.2cm}
    \caption{Accuracy for other motion or solver parameters. (a) shows the error of the solution over an increasing scale of the velocity. (b) indicates the effect of the time interval length. (c) shows the effect of the number of observed lines.}
\end{figure}

To evaluate the proposed solver, we adopt the single variable method to conduct various simulation experiments from two aspects. One is to evaluate the solver's robustness against noise including disturbance of events, errors of the extracted lines at the boundary of the interval {and disturbances of angular velocities}. 
The other one is to investigate the effect of certain factor such as the scale of the velocity, the length of the time interval, or the number of lines. Note that the errors for each level of each variable are averaged over 500 experiments. The detail configurations are as follows:
\begin{itemize}
    \item  \textbf{Robustness against event location noise:} The disturbance of each event is varied with a standard deviation reaching from 0 to 5 pixels with a step size of 0.5 pixels. The average $\mu$ and standard deviation $\sigma$ of $\epsilon$ and $\phi$ is presented in Figure~\ref{fig:eventline_eventNoise}. 
    \item  \textbf{Robustness against noise in the end points of $l_{1j}$ and $l_{3j}$:} We add pixel-level Gaussian noise to the two endpoints of the projected lines to test the robustness of the proposed linear solver. The noise is varied between 0 and 5 pixels with a step size of 0.5 pixels. Figure~\ref{fig:eventline_lineNoise} indicates the simulated results.
    {
    \item  \textbf{Robustness against noise in angular velocity:}  We add Gaussian noise to the known angular velocity to test the robustness of the proposed linear solver. The noise is varied between 0 and 1 rad/s with a step size of 0.1 rad/s. Figure~\ref{fig:eventline_angularNoise} indicates the respective results.}
    \item  \textbf{Effect of speed:}  We set the direction of the linear velocity as $[ 0.447,\ 0.894,\ 0]$ and the scale of the velocity is varied between 0 and 10 m/s. The simulation results are shown in Figure~\ref{fig:eventline_speed}. As can be observed, errors are decreasing with an increasing norm of the speed. In other words, the higher speed, the more accurate our solver is operating.
    \item \textbf{Effect of the time interval size:} We vary the time interval from 0.2 s to 2.2s with a step size of 0.2 s. Results (Figure~\ref{fig:eventline_timeInterval}) indicate that the errors are decreasing as the time interval is increasing.
    \item \textbf{Effect of the number of lines:} The number of lines is varied from 2 to 10 with steps of 1. Figure~\ref{fig:eventline_lineNum} presents the results. The more lines are present in the scene, the higher the accuracy of the solver.
\end{itemize}

Without loss of generality, the errors increase as noise level are increasing (cf. Figure~\ref{fig:eventline_noise}). Note that the solver is rather sensitive to noise, which is analogous to the trifocal tensor-based approaches for standard cameras~\cite{hartley1997lines,weng1992motion}. Furthermore, the more displacement the camera experiences during the chosen time interval, the higher the expected accuracy.

%



%
%

\subsection{Real Data}

To the best of our knowledge, we are the first to propose the CELC-based linear velocity solver for event cameras, and as such it is hard to compare against an existing SOTA algorithm. We therefore design our own baseline algorithm to evaluate the proposed method, which is based on the line-line-line incidence relationship~\cite{hartley2003multiple}.
\begin{equation}
    \mathbf{l}_2 \times (\mathbf{l}_1^{\mathsf{T}}[\mathbf{T}_1,\mathbf{T}_2,\mathbf{T}_3]\mathbf{l}_3) = 0.
\end{equation}
Using equation~(\ref{eqn:eventline_CELC}), the line-line-line constraint under continuous motion is given by
\begin{equation}
    \mathbf{l}_2^{\wedge} \mathbf{B}_k \mathbf{v} = 0,
\end{equation}
where $\mathbf{l}_2^{\wedge}$ is the $3\times3$ screw symmetric matrix form of $\mathbf{l}_2$. The additional line $l_{2}$ is fitted at the center of the interval and by using the same strategy as introduced in Sec~\ref{sec:line_extraction}. With line features $\mathbf{l}_1, \mathbf{l}_2, \mathbf{l}_3$ in three views as well as known angular velocity, we can again stack the constraints for all clusters and figure out the linear velocity through a similar robust nullspace calculation method as before. We denote this baseline implementation the continuous event-based line-line-line constraint~(CE3LC). 

{We verify feasibility and practicality of our approach on two real-world datasets collected by an automated guided vehicle (AGV) and an unmanned aerial vehicle (UAV), respectively. The two datasets are collected by a DAVIS346, which has a resolution of 346x260 pixels.}

\vspace{0.2cm}
\noindent\textbf{(1) AGV with a downward-facing event camera}

\begin{figure}[t]
    \centering
    \subfigure[]{
        \includegraphics[width = 0.23\textwidth]{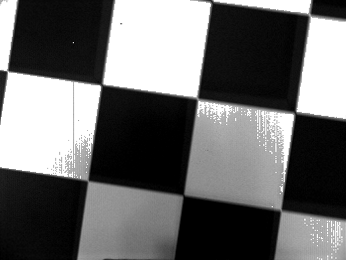}
        \label{fig:mplImg}
    }
    \subfigure[]{
        \includegraphics[width = 0.24\textwidth]{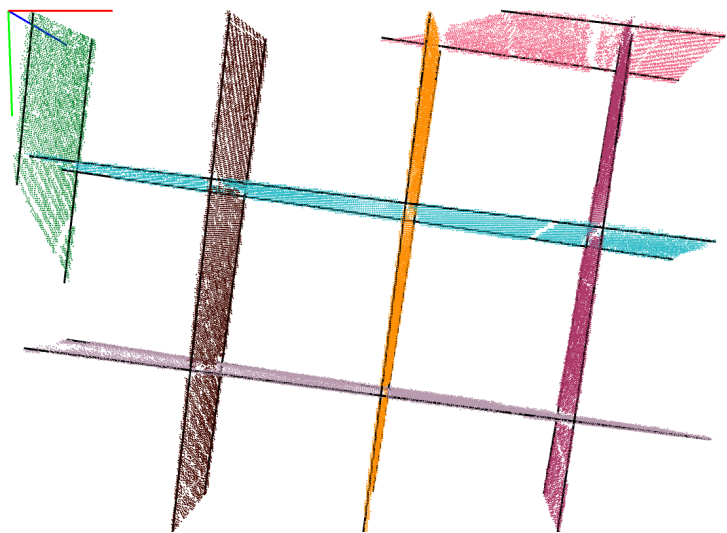}
        \label{fig:mplCluster}
    }
    \subfigure[]{
        \includegraphics[width = 0.23\textwidth]{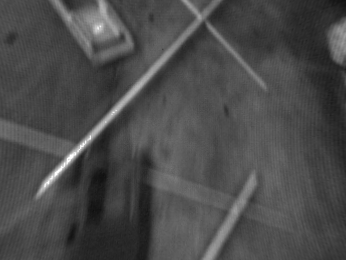}
        \label{fig:fpvImg}
    }
    \subfigure[]{
        \includegraphics[width = 0.215\textwidth]{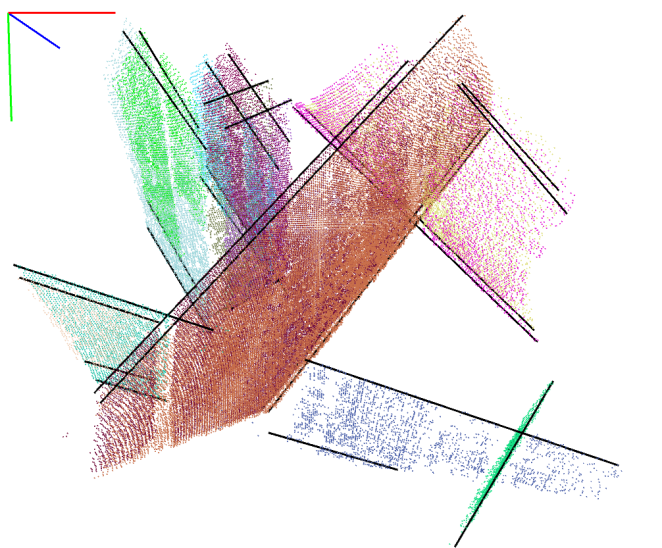}
        \label{fig:fpvCluster}
    }
    \caption{(a)-(b) Example data captured by an AGV with a downward-facing event camera. (c)-(d) Example data captured by an UAV with a $45^{\circ}$ downward-facing event camera. (a) and (c) denote grayscale images, whereas (b) and (d) the identified event clusters corresponding to real-world line segments in a spatio-temporal view. The extracted lines at the beginning and at the end of each interval are shown in black. The coordinate system in the upper left corner of (b) and (d) means uses the red and green axes to denote the x and y coordinates of each event, and the blue axis to indicate the temporal axis.}
\end{figure}

The AGV dataset is collected with a camera mounted on the front of an XQ-4 Pro robot and faces downwards (see Figure \ref{fig:mplImg}-\ref{fig:mplCluster}). We recorded a uniform circular motion sequence on a chessboard. Ground truth is obtained via an Optitrack motion capture system. Our algorithm is working in normalized coordinates, which is why normalization and undistortion are computed in advance.

Figure~\ref{fig:mplCluster} shows the collected data and line clusterings. To alleviate the influence of noise on the events and resulting inaccuracies in the line clusters and extraction, we utilize a spatio-temporal window with 1,000,000 events (about 0.7s) to estimate the linear velocity.

\vspace{0.2cm}
\noindent \textbf{(2) UAV with a $45^{\circ}$ downward-facing event camera}


We further evaluate our method on a sequence (Indoor45 9) from \cite{Delmerico19icra} which is captured by an UAV equipped with a $45^{\circ}$ downward-facing event camera with a resolution of 346$\times$260 pixels. The maximum velocity ($|\overset{\rightarrow}{v}|_{max}$) of the UAV is about 11.23 $m/s$, which means the intensity images are rather blurry. Therefore, it is difficult to use the intensity images to extract lines for pose estimation. The event camera however works well in such a challenging scenario. As can be observed in Figure~\ref{fig:fpvCluster}, our strategy maintains successfully extracted event clusters and starting and ending lines from the raw stream of events. Note that in order to better distinguish lines that are very close, we separate events into positive and negative sets before running the actual clustering algorithm.


\subsubsection{Analysis of the Results}

We select 5 sequences from each dataset and the results are listed in Table~\ref{tab:agv_errors} and Table~\ref{tab:uav_errors}. CELC indicates the proposed solver without optimization, CELC+opt the proposed solver with nonlinear optimization, and CE3LC the proposed baseline algorithm {without nonlinear optimization}. As can be observed, CELC+opt typically outputs better results than CELC, indicating the positive impact of nonlinear optimization. Furthermore, CELC outperforms CE3LC. The reason is given by the fact that CE3LC relies more heavily on the performance of 2D line fitting, while CELC utilizes all events measurements to constrain the problem.

Note that the accuracy of the algorithm highly depends on the accuracy of the line detection and fitting, the resolution of the camera, the number of lines in the environment, and other factors analyzed in the simulation experiments. For example, the resolution we use is only 346$\times$260 pixels. As demonstrated by the KITTI dataset \cite{Geiger2012CVPR}, a common resolution for normal cameras would be in the order of 1392$\times$512, which is much higher. We furthermore believe that the number of studies on line detection and fitting in event streams is still rather limited, and better approaches would certainly benefit the method proposed in this paper.

\begin{table}
\centering
    \begin{minipage}[t!]{0.48\textwidth}
    \centering
    \caption{AGV Errors}
    \label{tab:agv_errors}
    \resizebox{\textwidth}{!}{
        \begin{tabular}{cc|cccc}
        \toprule
        \multicolumn{2}{c}{\textbf{Method}} & CELC & CELC+opt & CE3LC \\ \midrule
        \multirow{2}{*}{\textbf{Seq1}}  & $\epsilon$ [m/s]   & 0.2058      & \textbf{0.2035}   & 0.6345            \\ 
                                        & $\phi$ [rad]       & 0.2063     & \textbf{0.2038}  & 0.6457   \\ 
                                        \hline
        \multirow{2}{*}{\textbf{Seq2}}  & $\epsilon$ [m/s]   &\textbf{0.1125}      &0.1204    &0.3180            \\ 
                                        & $\phi$ [rad]       &\textbf{0.1123}     &0.1201    &0.3192             \\ 
                                        \hline
        \multirow{2}{*}{\textbf{Seq3}}  & $\epsilon$ [m/s]   &0.2042      &\textbf{0.1476}   &0.6783            \\ 
                                        & $\phi$ [rad]       & 0.2043      &\textbf{0.1471}    & 0.6921            \\ 
                                        \hline
        \multirow{2}{*}{\textbf{Seq4}}  & $\epsilon$ [m/s]   & 0.1590     &\textbf{0.1455}    & 0.1951            \\ .
                                        & $\phi$ [rad]       &0.1589      & \textbf{0.1450}   &  0.1952           \\ 
                                        \hline
        \multirow{2}{*}{\textbf{Seq5}}  & $\epsilon$ [m/s]   &0.2149      &\textbf{0.1439}    &1.0122             \\ 
                                        & $\phi$ [rad]       &0.2154      &\textbf{0.1441}    &1.0615             \\
        \bottomrule
        \end{tabular}
        }
    \end{minipage}
    \begin{minipage}[t!]{0.48\textwidth}
    \centering
    \caption{UAV Errors}
    \label{tab:uav_errors}
    \resizebox{\textwidth}{!}{
        \begin{tabular}{cc|cccc}
        \toprule
        \multicolumn{2}{c}{\textbf{Method}} & CELC & CELC+opt & CE3LC \\ \midrule
        \multirow{2}{*}{\textbf{Seq1}}  
        & $\epsilon$ [m/s] & \textbf{0.2145} & 0.2187 & 0.5192 \\ 
        & $\phi$ [rad] &\textbf{0.2150} & 0.2193 & 0.5326 \\ 
        \hline
        \multirow{2}{*}{\textbf{Seq2}}  
        & $\epsilon$ [m/s] & 0.2062 & \textbf{0.1936} & 0.6263 \\
        & $\phi$ [rad] & 
        0.2067 & \textbf{0.1940} & 0.6536 \\
        \hline
        \multirow{2}{*}{\textbf{Seq3}}  
        & $\epsilon$ [m/s] & 0.3619 & \textbf{0.2499} & 0.4756 \\ 
        & $\phi$ [rad] &0.3661 & \textbf{0.2507} & 0.4922 \\
        \hline
        \multirow{2}{*}{\textbf{Seq4}}  
        & $\epsilon$ [m/s] & 0.2340 & \textbf{0.2108} & 0.5297 \\ 
        & $\phi$ [rad] & 0.2347 & \textbf{0.2110} & 0.5379 \\
        \hline
        \multirow{2}{*}{\textbf{Seq5}}  
        & $\epsilon$ [m/s] & 0.2126 & \textbf{0.1118} & 0.4828 \\ 
        & $\phi$ [rad] & 0.2138 & \textbf{0.1119} & 0.4924 \\
        \bottomrule
        \end{tabular}
        }
    \end{minipage}%
\end{table}


\section{CONCLUSION}
\label{sec:eventline_conclusion}

Different from existing event-based motion estimation approaches, we are the first to exploit trifocal tensor geometry in order to constrain the dynamics of an event camera from an event stream generated by the continuous observation of arbitrary 3D lines. The closed-form velocity solver employs a novel constraint which we denote the Continuous Event Line Constraint (CELC). We believe that our algorithm is an important first step into the direction of velocity bootstrapping for DVS sensors, and our future work considers the embedding of this solver and the related constraints into a more complete, event-based visual-inertial framework for direct velocity estimation.

\bibliography{egbib}
\end{document}